\documentclass[letterpaper, 10 pt, conference]{ieeeconf}  % Comment this line out if you need a4paper

\IEEEoverridecommandlockouts                              % This command is only needed if
\usepackage{graphics} % for pdf, bitmapped graphics files
\usepackage{epsfig} % for postscript graphics files
\usepackage{mathptmx} % assumes new font selection scheme installed
\usepackage{newtxtext,newtxmath}
\usepackage{times} % assumes new font selection scheme installed
\usepackage{amsmath} % assumes amsmath package installed
\usepackage{amssymb}  % assumes amsmath package installed
\usepackage{lipsum}
\usepackage{mathtools}
\usepackage{cuted}
\usepackage{arydshln}
\usepackage{subcaption}
\usepackage{hyperref}

\title{\LARGE \bf Upset Recovery Control for Quadrotors Subjected to a Complete Rotor Failure from Large Initial Disturbances 
}

\author{Sihao Sun*, Matthias Baert*, Bram Strack van Schijndel and Coen de Visser% <-this % stops a space
 \thanks{*These two authors contributed equally}
\thanks{The authors are with Faculty of Aerospace Engineering, Delft University of Technology, 2629 HS Delft, The Netherlands.
        {email: s.sun-4@tudelft.nl; c.c.devisser@tudelft.nl}}
\thanks{The video is available at:\textbf{ https://youtu.be/PJ5U3ZAm8NM}}
}

\begin{document}
\maketitle
%\thispagestyle{empty}
%\pagestyle{empty}

%%%%%%%%%%%%%%%%%%%%%%%%%%%%%%%%%%%%%%%%%%%%%%%%%%%%%%%%%%%%%%%%%%%%%%%%%%%%%%%%
\begin{abstract}
This study has developed a fault-tolerant controller that is able
to recover a quadrotor from arbitrary initial orientations and angular velocities, despite the complete failure of a rotor. This cascaded control method includes a position/altitude controller, an almost-global convergence attitude controller, and a control allocation method based on quadratic programming. As a major novelty, a constraint of undesirable angular velocity is derived and fused into the control allocator, which significantly improves the recovery performance. For validation, we have conducted a set of Monte-Carlo simulation to test the reliability of the proposed method of recovering the quadrotor from arbitrary initial attitude/rate conditions. In addition, real-life flight tests have been performed. The results demonstrate that the post-failure quadrotor can recover after being casually tossed into the air.
% With only 3 rotors the quadrotor is
% not fully controllable anymore, however by using precession the
% quadrotor can be made controllable again in roll and pitch. A
% step-by-step recovery strategy is proposed in which controllability
% is actively recovered before recovery of attitude and altitude.
% Control allocation is based on the Incremental Non-Linear
% Dynamic Inversion (INDI) principle and takes rotor saturation
% into account by solving a constrained quadratic optimization
% problem. The controller is validated in a real-life test environment
% where the quadrotor is thrown into the air with only 3 propellers
% from which it has to recover using the techniques presented in
% this paper.
\end{abstract}
% \begin{IEEEkeywords}Robot Safety; Aerial Systems: Mechanics and Control
% \end{IEEEkeywords} 

%%%%%%%%%%%%%%%%%%%%%%%%%%%%%%%%%%%%%%%%%%%%%%%%%%%%%%%%%%%%%%%%%%%%%%%%%%%%%%%%
\section{Introduction}
In recent years, multi-rotor aerial vehicles have received a
lot of attention. These aerial vehicles are usually unmanned
robots that can perform various tasks, in some cases without
human intervention. Multi-rotors are mainly used outdoors for
agricultural purposes, architecture and construction, delivery,
emergency services, media purposes or to monitor and conserve
the environment. As these vehicles will become more
involved in daily life, safety can not be overlooked.

One of the most common multi-rotors is the quadrotor
due to its simplicity and energy efficiency~\cite{Stolaroff2018}. As the name
implies, a quadrotor has four rotors positioned in a rectangular
profile on the vehicle. However, because this vehicle is not
over-actuated, this type of multi-rotor suffers most from an
actuator failure and might not be able to continue its mission
or worse, might not be able to land safely.

\subsection{Fault-Tolerant Control}
Fault-tolerant control (FTC) for quadrotors has been the subject
of various literature sources. Some research is focused on the
partial damage of a rotor \cite{Li2013,Wang2019}, while other research considers
the complete loss of one or multiple rotors. 
A solution to the case of a complete loss of a rotor is
presented in \cite{Lanzon2014} where the author proposes to give
up on yaw control to maintain control over the other
states. An analytical solution
under the complete loss of one, two or three propellers are given in \cite{Mueller2014,Mueller2015}.
A PID and a backstepping approach
focusing on an emergency landing in case of failure is presented in \cite{Lippiello2014,Lippiello2014a} respectively. A fault-tolerant controller using incremental nonlinear dynamic inversion (INDI) is given in \cite{Lu2015} where fault detection is also implemented. To improve the robustness of the controller, \cite{Hou2020} employs a nonsingular terminal sliding mode control (NTSMC) to this fault-tolerant control problem.

The validations in practice are carried out by \cite{Mueller2014} using the linear quadratic regulator (LQR) around the proposed analytical equilibrium. To improve the stability under various yaw rates, the study in \cite{Stephan2018} employs a linear parameter varying (LPV) controller. 
In \cite{Sun2018}, a quadrotor with loss of single rotor controlled by INDI is shown able to fly in high-speed conditions despite significant aerodynamic disturbances. 
% The research does not take rotor
% saturation into account. This means that at high speeds (9
% m/s) the quadcopter loses control as the rotors start saturating. In \cite{Smeur2017} an improved INDI control allocation method
% is presented which takes rotor saturation into account.

\begin{figure}
    \centering
    \includegraphics[scale=0.4]{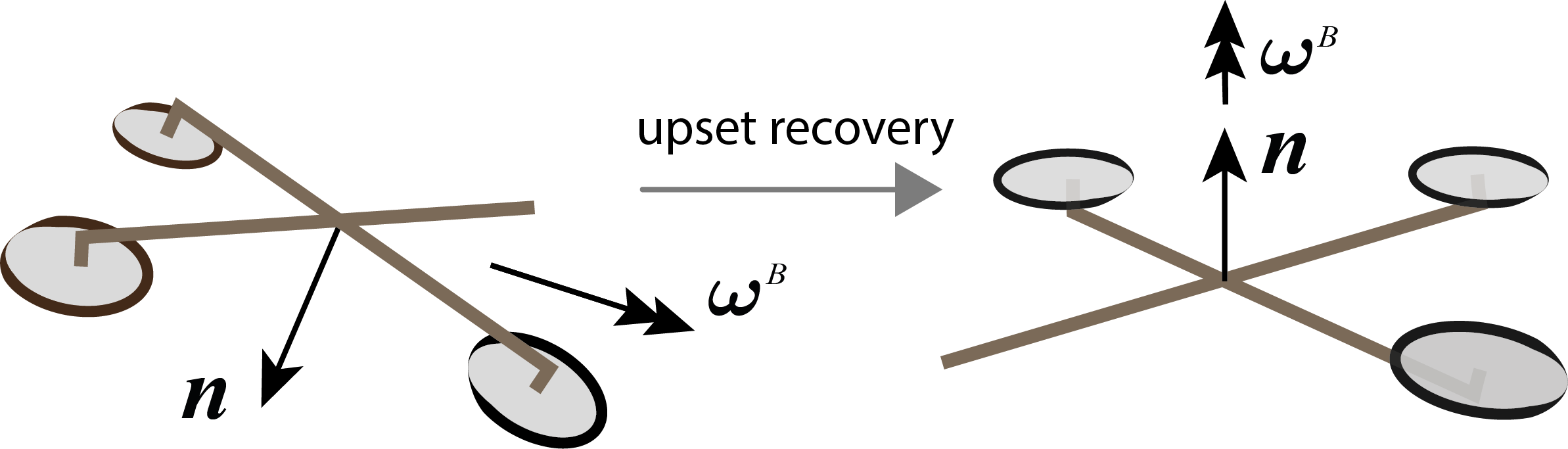}
    \caption{Illustration of the upset recovery problem where $\boldsymbol{n}$ indicates the total thrust direction and $\boldsymbol{\omega}^B$ indicates the vehicle angular velocity.}
    \label{fig:recovery_illustration}
\end{figure}

% \textit{Upset recovery} is a technique extensively studied for improving aviation safety~\cite{Crespo2012,Engelbrecht2013}. The upset condition is defined as "any uncommanded or inadvertent event with an abnormal aircraft attitude, angular rate, acceleration, airspeed, or flight trajectory"~\cite{}. Compared to an aircraft, a multi-rotor drone rarely experience upset conditions thanks to high control effectiveness. It can simply perform aerobatic maneuvers such as somersaults. This, however, has to be considered for quadrotors with rotor failures due to the significant reduction of its maneuverability. 

% . Unfortunately, such large disturbances with respect to the hovering condition can be very likely to happen, because the rotor failure could occur during aggressive maneuvers, after air collision or even due to the unexpected long fault detection time.
% % To the best of our knowledge, none of the existing fault-tolerant control method has the ability to recover the faulty drone from these upset conditions.
\subsection{Upset Recovery}
Upset recovery is a technique extensively studied for improving aviation safety~\cite{Crespo2012,Engelbrecht2013}. The upset condition is defined as "any uncommanded or inadvertent event with an abnormal aircraft attitude, angular rate, acceleration, airspeed, or flight trajectory~\cite{Belcastro2011}", such as aircraft stall that directly leads to loss-of-control~\cite{Lambregts2008}. In comparison, upset of a multi-rotor drone is rarely heard by virtue of its relatively high control effectiveness in full flight envelope. For example, a quadrotor can easily perform aerobatic maneuvers~\cite{Faessler2015}.

However, due to the significant maneuverability reduction, a quadrotor with single-rotor-failure can easily enter an upset condition. For instance, as Fig.~\ref{fig:recovery_illustration} shows, a post-failure quadrotor may be upside down and fast rotating before the FTC is triggered, because of strong wind disturbances and delay of the fault detection module. At this moment, existing FTC methods could fail owing to multiple reasons, such as violation of linearization assumptions, actuator saturations, etc. Therefore, an improved FTC method is required to address the upset recovery problem.

\subsection{Contributions}
As the main contribution, this research proposes a controller which has the ability to recover a quadrotor with complete loss of a rotor from an arbitrary attitude and a wide range of initial angular velocities. Then the method can subsequently steer the damaged drone to a designated position and altitude. This cascaded control method is composed of three parts: a control allocator that tracks the angular acceleration command while suppressing the undesirable angular rate, an attitude controller with an almost-global attraction region, and a position controller subordinate to the former two parts. 

The control method has been validated in a real-life environment where the quadrotor was randomly tossed into the air and recovers thereafter. A set of Monte-Carlo simulations have been also performed to test the performance of the controller from a wide range of initial conditions. It is shown that the proposed method can significantly improve the quadrotor safety after rotor failures despite large initial disturbances.

% Recovery or more specific upset recovery is rarely considered
% in quadrotor research. In Faessler et al. the recovery of
% a quadrotor is discussed, but the focus is on the recovery of
% the vision-based state estimation [14]. The quadrotor is thrown
% into the air by hand in order to create an upset condition. For
% airplanes upset recovery is a more common field of study.
% Many literature sources propose a step-by-step recovery process
% when an airplane is in an upset condition [15, 16]. Plenty
% of literature sources cover trajectory generation and control
% for quadrotors, which is also related to recovery [17, 18, 19].
% Hehn and Andrea present a trajectory generation solution, that
% should be able to solve the recovery problem in an abstract
% way [20].

% Recovery of a quadrotor with a rotor failure, however, has not been addressed in any previous research. In such an abnormal condition, the quadrotor has much less maneuverability for attitude control. For this reason, even with the state-of-art fault-tolerant control, the quadrotor could still 

% In Section II the model of the quadrotor used is explained.
% Section III presents the proposed fault-tolerant control system
% that is able to recover from an upset condition. In Section IV
% the experimental setup is explained while in Section V the
% results are presented. The paper concludes with Section VI.
\section{Problem Formulation}
\subsection{Notation}
The inertial frame is represented by the north-east-down coordinate system. The body frame is originated at the c.g. of the vehicle with the forward-right-down convention, as is shown in Fig.~\ref{fig:quadrotor_illustration}. Throughout the paper, we use lower-case boldface symbols to denote vectors, upper-case boldface symbols for matrices and non-boldface symbols for scalars. A 3-D vector with superscript $[\cdot]^B$ indicates that the vector is expressed in the body frame, otherwise in the inertial frame. Operator $\mathrm{diag}(\cdot)$ indicates a diagonal matrix with element ($\cdot$) as diagonal entries.
\subsection{6-DoF Model of a Quadrotor}
\begin{figure}
    \centering
    \includegraphics[scale=0.45]{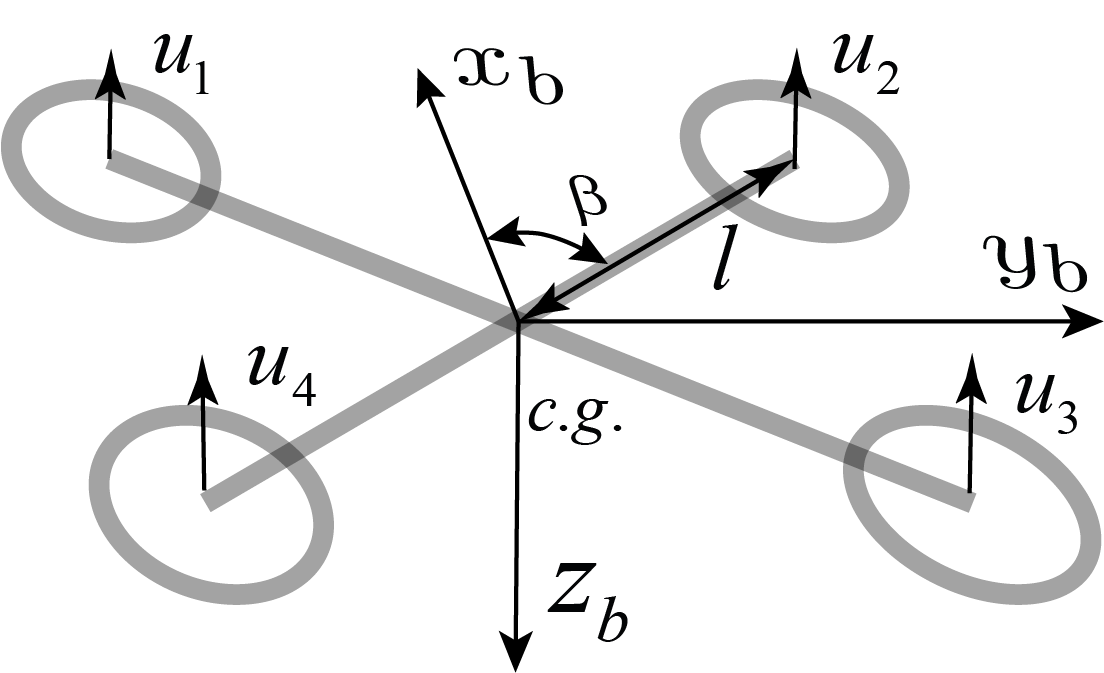}
    \caption{Definition of the body frame, the geometric parameters $\beta$ and $l$, the index of control inputs.}
    \label{fig:quadrotor_illustration}
\end{figure}
The quadrotor is powered by four independently controlled rotors to produce necessary lift and control moments. Fig.~\ref{fig:quadrotor_illustration} shows the definition of the body frame, and the rotor index of a quadrotor. The state equations of a quadrotors can be composed of the following 6-DoF rigid body kinematics and dynamic equations \cite{Mahony2012}:
\begin{equation}
    \dot{\boldsymbol{\xi}} = \boldsymbol{v}
\end{equation}
\begin{equation}
    \boldsymbol{\dot{R}} =\boldsymbol{R} \boldsymbol{\omega}_\times^B
\end{equation}
% \begin{equation}
%     \dot{\boldsymbol{v}}^B+\boldsymbol{\omega}^B\times \boldsymbol{v}^B = \boldsymbol{L}_{BI}\boldsymbol{g} + (\boldsymbol{f}_c^B+\boldsymbol{f}_{a}^B)/m
%     \label{eq:F_dyn_equation}
% \end{equation}
\begin{equation}
    m\dot{\boldsymbol{v}}= m\boldsymbol{g} + \boldsymbol{R}(\boldsymbol{f}_c^B+\boldsymbol{f}_{a}^B)
    \label{eq:F_dyn_equation}
\end{equation}
\begin{equation}
    \boldsymbol{I}_v^B\dot{\boldsymbol{\omega}}^B + \boldsymbol{\omega}^B\times \boldsymbol{I}_v^B\boldsymbol{\omega}^B=  \boldsymbol{m}_c^B + \boldsymbol{m}_a^B + \boldsymbol{m}_g^B
    \label{eq:M_dyn_equation}
\end{equation}
where $\boldsymbol{\xi} = [x,~y,~z]^T$ and $\boldsymbol{v} = [v_x,~v_y,~v_z]^T$ indicate the position and velocity respectively. $\boldsymbol{R} \in$ SO(3) is the rotational matrix of the quadrotor from the body frame to the inertial frame. Therefore, for any vector $\boldsymbol{e}\in \mathbb{R}^3$, we have $\boldsymbol{e} = \boldsymbol{R}\boldsymbol{e}^B$. The angular velocity of the body frame w.r.t the inertial frame is expressed as $\boldsymbol{\omega}^B = [\omega_x,~\omega_y,~\omega_z]^T$, where $\boldsymbol{\omega}_{\times}^B$ is the skew symetric matrix such that $\boldsymbol{\omega}_{\times} \boldsymbol{a} = \boldsymbol{\omega}\times \boldsymbol{a}$ for any $\boldsymbol{a}\in \mathbb{R}^3$. The vehicle mass and inertia are denoted by $m$ and $\boldsymbol{I_v}^B$ respectively and $\boldsymbol{g}$ denotes the gravity vector. The control forces $\boldsymbol{f}_c^B$ and moments $\boldsymbol{m}_c^B$ are produced by rotors. A simplified model of forces and moments generated by rotors are expressed as
\begin{equation}
\boldsymbol{f}_c^B = [0,~0,~\boldsymbol{G}_t\boldsymbol{u}]^T,~~~\boldsymbol{m}_c^B = \boldsymbol{G}_m\boldsymbol{u} 
\label{eq:force_model_simple}
\end{equation}
% \begin{equation}

% \end{equation}
where $\boldsymbol{u} = [u_1,~u_2,~u_3,~u_4]^T$ and $u_i$ is the force produced by rotor $i$ (see Fig.~\ref{fig:quadrotor_illustration}). Note that $0\leq\boldsymbol{u}_\mathrm{min}\leq\boldsymbol{u}\leq\boldsymbol{u}_\mathrm{max}$. When complete failure of rotor $i$ occurs, we have $u_{\mathrm{min},i} = u_{\mathrm{max},i} = 0$. $\boldsymbol{G}_m$ is a mapping from rotor generated forces to control moments and $\boldsymbol{G}_t$ is the mapping from rotor generated forces to the total thrust. For a quadtrotor that the thrust of each rotor is parallel with the $\boldsymbol{z}_b$ axis, we have
\begin{equation}
    \boldsymbol{G}_t = [-1,-1,-1,-1]
\end{equation}
\begin{equation}
    \boldsymbol{G}_m = \mathrm{diag}(l\sin\beta,~l\cos\beta,~s\sigma)\left[
    \begin{array}{cccc}
         1& -1& -1 & 1  \\
         1& 1& -1 & -1  \\
         1& -1& 1 & -1  \\
    \end{array}
    \right]
\end{equation}
where $l$ and $\beta$ are geometric parameters as shown in Fig.~\ref{fig:quadrotor_illustration}. $\sigma$ is the torque thrust ratio of the rotor, $s$ is a sign variable determined by the rotating direction of the rotor. 

The force model given by (\ref{eq:force_model_simple}) neglects the variation of thrust stem from quadrotor translational motions with respect to the airflow. Therefore, an aerodynamic force term $\boldsymbol{f}_a^B$ is added in (\ref{eq:F_dyn_equation}), so as the term $\boldsymbol{m}_a^B$ in (\ref{eq:M_dyn_equation}). The gyroscopic moment, denoted by $\boldsymbol{m}_g^B$, is caused by the rotation of rotors with respect to the body frame.
% , which can be expressed as
% \begin{equation}
%     \boldsymbol{m}_g^B = -\boldsymbol{\omega}^B\times\sum{\boldsymbol{I}_p^B\boldsymbol{\omega}_i^B}-\sum{\boldsymbol{I}_p^B\dot{\boldsymbol{\omega}}_i^B}
% \end{equation}
For the current research, we omit $\boldsymbol{f}_a^B$,~$\boldsymbol{m}_a^B$ and $\boldsymbol{m}_g^B$ in the controller design whereas they are included in the simulation presented in Sec.\ref{sec:simulation}.

\subsection{Quadrotor Upset Recovery Problem}
A quadrotor has four independently powered rotors, such that the thrust, pitch, roll and yaw channels can be totally decoupled. This characteristic, however, can be different when a single rotor failure occurs. A most commonly used strategy is by giving up the yaw control and keep the rest which is more essential for maintaining the desired position and altitude \cite{Lanzon2014}. This requires the post-failure vehicle to enter a so-called \textit{relaxed-hovering condition}
\cite{Mueller2015} in which the drone spins about an average thrust direction whilst the position of the spinning center and the altitude maintain constant. By slightly changing the direction and amount of the reference thrust, the average position and altitude of the post-failure quadrotor can be controlled. 

Driving a quadrotor with a single rotor failure to the relaxed-hovering condition from arbitrary initial attitude, angular rates, and positions, is defined as the quadrotor upset recovery problem. 

% This can be further divided into three subproblems. Define the current thrust orientation as $\boldsymbol{n}$, the first subproblem is to control the orientation of thrust to the desired upward direction. The second subproblem is to regulate the angular rate of $\boldsymbol{\omega}^B$ to the angular rate during the relaxed-hovering condition. The third problem is controlling the altitude and position after the attitude and rates are recovered.

% The major challenge of the problem is twofold. First of all, an 
\section{Methodology}
\label{sec:methodology}
\begin{figure}
    \centering
    \includegraphics[scale=0.25]{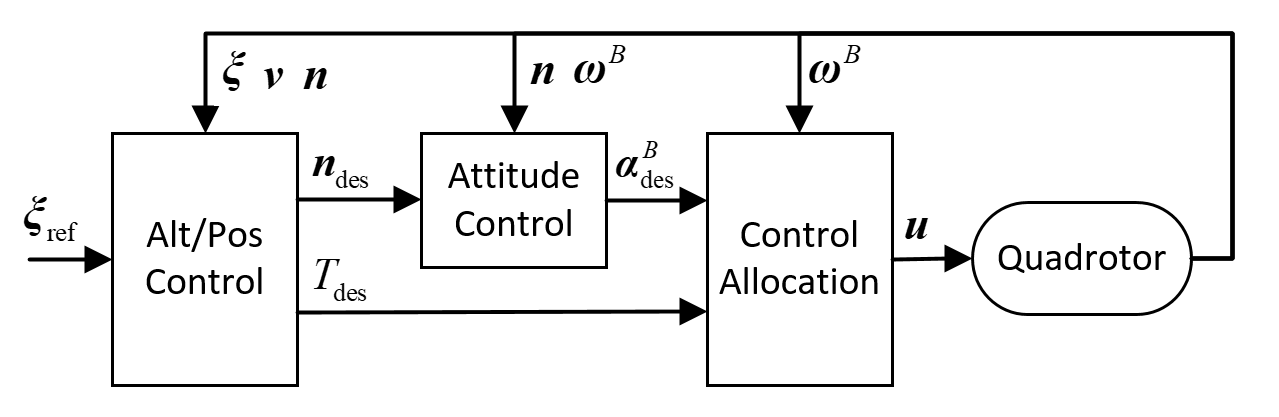}
    \caption{Diagram of the proposed control method.}
    \label{fig:control_diagram}
\end{figure}

The major challenge of the recovery problem is threefold. First of all, we need to design an almost-global (excluding finite singularities) reduced attitude controller to drive the vehicle orientation to the relaxed-hovering condition from large initial attitude deviation. Secondly, with the complete failure of a rotor, the quadrotor system only has three remaining constraint inputs. Hence it requires a novel control allocation approach to address input constraints while preventing the drone from entering upset conditions. Last but not least, a hedging of position/altitude loop need to be designed to coordinate with the aforementioned attitude controller and the control allocation method. A cascaded framework of the proposed controller is given as Fig.~\ref{fig:control_diagram} shows.
\subsection{Altitude and Position Control}
The position and altitude control, namely the outer-loop control, is designed as a cascaded P+PI controller as follows
\begin{equation}
    \boldsymbol{v}_\mathrm{des} = \boldsymbol{K}_{p,\mathrm{pos}}(\boldsymbol{\xi}_\mathrm{ref}-\boldsymbol{\xi})
\end{equation}
\begin{equation}
    \boldsymbol{a}_\mathrm{des,0} = \boldsymbol{K}_{p,\mathrm{vel}}(\boldsymbol{v}_\mathrm{des}-\boldsymbol{v}) + \boldsymbol{K}_{i,\mathrm{vel}}\int(\boldsymbol{v}_\mathrm{des}-\boldsymbol{v})\mathrm{d}t - \boldsymbol{g}
\end{equation}
where $\boldsymbol{\xi}_\mathrm{ref}$ is the reference position; $\boldsymbol{K}_{p,\mathrm{pos}}$, $\boldsymbol{K}_{p,\mathrm{vel}}$ and $\boldsymbol{K}_{i,\mathrm{vel}}$ are 3$\times$3 positive diagonal gain matrices.
The acceleration reference is then obtained by
\begin{equation}
    \boldsymbol{a}_\mathrm{des} = \mathrm{diag}(1/\epsilon,~1/\epsilon,~1)\boldsymbol{a}_\mathrm{des,0}
    \label{eq: a_des_constraint}
\end{equation}
where 
% \begin{equation}
% \epsilon = \max\left(\frac{\sqrt{a_{x,\mathrm{des,0}}^2+a_{y,\mathrm{des,0}}^2}}{a_{z,\mathrm{des,0}}\tan\theta_1}~,~1\right)    
% \end{equation}
\begin{equation}
\epsilon = \max\left(\sqrt{a_{x,\mathrm{des,0}}^2+a_{y,\mathrm{des,0}}^2}/a_{z,\mathrm{des,0}}\tan\theta_1~,~1\right)    
\end{equation}
Then we can obtain the desired thrust direction
\begin{equation}
    \boldsymbol{n}_\mathrm{des} = \boldsymbol{a}_\mathrm{des}/||\boldsymbol{a}_\mathrm{des}||
\end{equation}
Note that the transform (\ref{eq: a_des_constraint}) guarantees that the angle between $\boldsymbol{n}_\mathrm{des}$ and the reverse of gravity $-\boldsymbol{g}$ is confined by angle $\theta_1$. Limiting this desired thrust direction can prevent aggressive spatial maneuvers during recovery.

% By limiting this desired thrust direction, or equivalently limiting the horizontal acceleration reference, we can effectively prevent the quadrotor from pushing itself back into an upset condition.

Now we use $\theta$ to denote the angle between current thrust direction $\boldsymbol{n}$ and $-\boldsymbol{g}$
\begin{equation}
    \theta = \arccos(-\boldsymbol{g}\cdot \boldsymbol{n}/||\boldsymbol{g}||)
\end{equation}
Then the original thrust command can be obtained by
\begin{equation}
    T_{\mathrm{des},0} = -m\cdot a_{z,\mathrm{des}}/\cos\theta
    \label{eq:az_body_frame_reference}
\end{equation}
% which is consequently feed into (\ref{eq:mu_ref_calculation}) to calculate the reference $\boldsymbol{\mu}_\mathrm{ref}$.
However, this method may deteriorate the attitude loop performance. Consider when the drone is upside down where $\theta \geq 90$~deg, (\ref{eq:az_body_frame_reference}) gives a negative thrust command; or when $\theta = 90$~deg, (\ref{eq:az_body_frame_reference}) leads to singularity. For this reason, a scaling factor $\beta$ is introduced which is scheduled by the total incline angle $\theta$, yielding
\begin{equation}
    \beta = \frac{\theta_2-\min(\max(\theta,~\theta_1),~\theta_2)}{\theta_2-\theta_1}
\end{equation}
where $\theta_1 < \theta_2$ are predetermined parameters (see Table.~\ref{tab:control_gains}). Finally the total thrust command is obtained by
\begin{equation}
    T_{\mathrm{des}} = -\beta \cdot m\cdot a_{z,\mathrm{des}}/\cos(\min(\theta,~\theta_1))
    \label{eq:az_body_frame_reference_final}    
\end{equation}
\subsection{Attitude Control}
The attitude controller calculates the the angular rate command in order to control the thrust orientation $\boldsymbol{n}$ to $\boldsymbol{n}_\mathrm{des}$. Now introduce the total incline angle $\rho$ as the angle from $\boldsymbol{n}_\mathrm{des}$ to $\boldsymbol{n}$, where
\begin{equation}
    \rho = \arccos(\boldsymbol{n}_\mathrm{des}\cdot\boldsymbol{n})
    \label{eq:theta_definition}
\end{equation}
Define the instant rotation vector $\boldsymbol{n}_c$ perpendicular to both $\boldsymbol{n}_\mathrm{des}$ and $\boldsymbol{n}$, we have
\begin{equation}
    \boldsymbol{n}_{c} = \boldsymbol{n}\times\boldsymbol{n}_\mathrm{des}/\sin\rho
    \label{eq:define_of_rotation_vector}
\end{equation}
The reference angular rate can be consequently obtained
\begin{equation}
    \boldsymbol{\omega}_\mathrm{des}^B = k_{p,\mathrm{att}}\cdot \rho \boldsymbol{R}^T\boldsymbol{n}_c
\end{equation}
where $k_{p,\mathrm{att}}$ is a positive gain.
Then the angular acceleration reference can be obtained by a proportional controller with a feed-forward term
\begin{equation}
    \boldsymbol{\alpha}_\mathrm{des}^B = \boldsymbol{K}_{p,\mathrm{rate}}(\boldsymbol{\omega}_\mathrm{des}^B-\boldsymbol{\omega}^B) + \dot{\boldsymbol{\omega}}_\mathrm{des}^B
\end{equation}
Note that (\ref{eq:define_of_rotation_vector}) becomes singular when $\boldsymbol{n}$ and $\boldsymbol{n}_\mathrm{des}$ are collinear. Thus the attitude control presented above could result in the almost-global reduced attitude stabilization~\cite{Fortescue2011} with exception of two special points, namely $\rho\in\{0,~\pi\}$. In practice, when singularity occurs, we can simply set $\boldsymbol{n}_c$ as an arbitrary unit vector perpendicular to $\boldsymbol{n}$.
% Since $\boldsymbol{n}^B = [0,~0,~-1]^T$ indicates the thrust direction, it is clear that the third element of $\boldsymbol{\omega}_{ref}$ should be zero at the ....
\subsection{Control Allocation}
The control allocation step solves the desired thrust of each rotor, namely $\boldsymbol{u}$, using the desired angular acceleration $\boldsymbol{\alpha}_\mathrm{des}^B$ and the total thrust command $T_\mathrm{des}$ as calculated above. Now, we use $\boldsymbol{\mu}_\mathrm{des}$ to denote the desired control moments and thrust. By replacing $\dot{\boldsymbol{\omega}}^B$ with $\boldsymbol{\alpha}_\mathrm{des}^B$ in (\ref{eq:M_dyn_equation}) and omitting $\boldsymbol{m}_g$ and $\boldsymbol{m}_a$, we have 
% \begin{equation}
%     \boldsymbol{\mu}_\mathrm{ref} = 
%     \left[\begin{array}{c}
%          \boldsymbol{m}_{c}\\
%          T
%     \end{array}\right]_{\mathrm{ref}} = \mathrm{diag}(\boldsymbol{I_v},~m)
%         \left[\begin{array}{c}
%          \boldsymbol{\alpha}_{\mathrm{ref}} \\
%          a_{z,{\mathrm{ref}} }
%     \end{array}\right] 
%     +
%     \left[\begin{array}{c}
%      \boldsymbol{\omega}\times \boldsymbol{I}_v\boldsymbol{\omega} \\
%     0
%     \end{array}\right] 
%     \label{eq:mu_ref_calculation}
% \end{equation}
\begin{equation}
    \boldsymbol{\mu}_\mathrm{des} = 
    \left[\begin{array}{c}
         \boldsymbol{m}_{c,\mathrm{des}}^B\\
         T_{\mathrm{des}}
    \end{array}\right] = 
        \left[\begin{array}{c}
         \boldsymbol{I}_v^B\boldsymbol{\alpha}_{\mathrm{des}}^B  + \boldsymbol{\omega}^B\times \boldsymbol{I}_v^B\boldsymbol{\omega}^B\\
         T_{\mathrm{des}}
    \end{array}\right] 
    \label{eq:mu_des_calculation}
\end{equation}
The thrusts generated by rotors need to cooperatively fulfil the reference represented by $\boldsymbol{\mu}_\mathrm{des}$. As the thrust produced by a rotor is limited, we establish a constrained Quadratic Programming (QP) problem to solve $\boldsymbol{u}$:
\begin{equation}
\begin{array}{cccc}
\mathrm{P}1:
    & &\underset{\boldsymbol{u}}{\text{min}}
    &\left(\boldsymbol{\mu}_{\mathrm{des}}-\boldsymbol{G}\boldsymbol{u}\right)^T\boldsymbol{W} \left(\boldsymbol{\mu}_{\mathrm{des}}-\boldsymbol{G}\boldsymbol{u}\right) + \lambda \boldsymbol{u}^T\boldsymbol{u}\\
    & &\text{s.t.} 
    & \boldsymbol{u}_\mathrm{min} \leq \boldsymbol{u} \leq \boldsymbol{u}_\mathrm{max}\\
\end{array}
\label{eq:P1_optimization}
\end{equation}
where $\boldsymbol{G} = [\boldsymbol{G}_m^T,~\boldsymbol{G}_t^T]^T$ is a combined control effective matrix;  $\boldsymbol{W} = \mathrm{diag}(W_x,~W_y,~W_z,~W_t)$ is a user defined weighting matrix, which determines the weight for each control objective; $\lambda>0$ is another weight for minimizing the control effort.

\begin{figure}
    \centering
    \includegraphics[scale = 0.45]{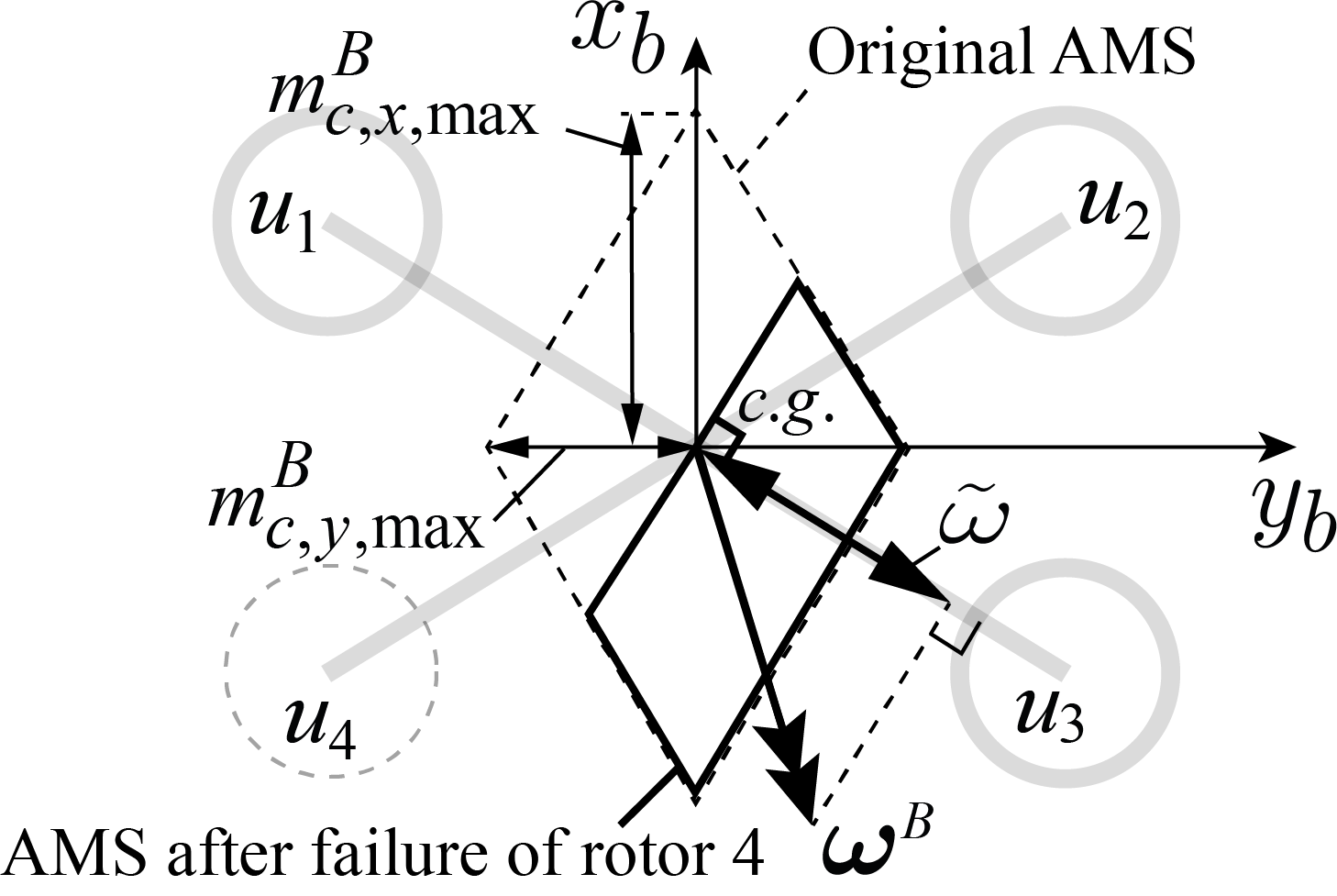}
    \caption{Projection of the attainable moment set (AMS) on the $x_b-y_b$ plane before and after the failure of rotor 4. The projection of current angular velocity $\boldsymbol{\omega}^B$ perpendicular to the boundary of AMS is unable to be reduced by the control moment. The magnitude of this component is denoted by $\tilde{\omega}$.}
    \label{fig:define_of_tilde_omega}
\end{figure}

P1 is a typical control allocation method for both aircraft and drones~\cite{Petersen2006,Smeur2017}. However, for a quadrotor with single rotor failure, we need to add an additional constraint to P1. We hereby define the \textit {Attainable Moment Set} (AMS) as a set of moments that can be generated by the existing rotors. As Fig.~\ref{fig:define_of_tilde_omega} shows, the area of AMS is reduced after rotor failure occurs. This is due to the fact that the quadrotor with fixed-pitch rotors can not generate negative lift, namely $\boldsymbol{u}_\mathrm{min}\geq 0$. In consequence, the angular velocity which cannot be suppressed by the current attainable moment will cause unstoppable rotations. The magnitude of this angular velocity, denoted by $\tilde{\omega}$, must be restrained during upset recovery.

% conducting allocation by solving P1 may cause rotations about an undesirable axis and is unable to effectively decelerate. This is due to the fact that the quadrotor with a fixed-pitch rotor can not generate negative lift, namely $\boldsymbol{u}_\mathrm{min}\geq 0$. We define the angular rate component on this \textit{unrecoverable rotation axis} as $\tilde{\omega}$, which is illustrated in Fig.~\ref{fig:define_of_tilde_omega}. In order to prevent the drone from entering this condition, $\tilde{\omega}$ must be constrained whilst performing control allocation. 

A constraint of $\tilde{\omega}$ after a brief time period $t_h$ is then introduced. Since the maneuverability on pitch/roll direction is much higher than yaw direction, we assume that $\omega_z$ in the period $t_h$ is constant. Recall (\ref{eq:M_dyn_equation}), and approximate $\boldsymbol{I}_v^B$ by $\mathrm{diag}(I_x,~I_y,~I_z)$, we have
\begin{equation}
         \left[\begin{array}{c}
         \dot \omega_x  \\
         \dot \omega_x 
    \end{array}\right]=\left[
    \begin{array}{cc}
        0 & \frac{I_y-I_z}{I_x}\omega_z \\
        \frac{I_z-I_x}{I_y}\omega_z & 0
    \end{array}\right]
    \left[
    \begin{array}{c}
         \omega_x  \\
         \omega_y 
    \end{array}
    \right]  \\
     +
     \hat{\boldsymbol{G}}_{m}\boldsymbol{u}
     \label{eq:wxwy_linear_equation}
\end{equation}
where 
\begin{equation}
    \hat{\boldsymbol{G}}_{m} = \mathrm{diag}(I_x,~I_y)^{-1}\left[
     \begin{array}{ccc}
         1 & 0 & 0  \\
         0 & 1 & 0
     \end{array}\right]
     \boldsymbol{G}_{m}
\end{equation}
Note that (\ref{eq:wxwy_linear_equation}) is a linear ODE, thus the time history of $\omega_x$ and $\omega_y$ can be explicitly solved with given initial conditions and control inputs. Assume the control input $\boldsymbol{u}$ is constant within $t_h$, and use the current $\omega_x$ and $\omega_y$ as initial conditions, then $\tilde{\omega}$ after $t_h$ can be expressed as
\begin{equation}
\Tilde{\omega}(t_h)=\boldsymbol{\phi}
    \left[\begin{array}{c}
         \omega_x(t_h)  \\
         \omega_y(t_h) 
    \end{array}\right]= \boldsymbol{\phi}\boldsymbol{\Phi}_0(t_h)
        \left[\begin{array}{c}
         \omega_x  \\
         \omega_y 
    \end{array}\right]+ \boldsymbol{\phi}\boldsymbol{\Phi}_1(t_h)\hat{\boldsymbol{G}}_{m}\boldsymbol{u} 
    \label{eq:tilde_omega_th}
\end{equation}
where $\boldsymbol{\phi}\in\mathbb{R}^{1\times2}$ is a row vector converting $\omega_x$ and $\omega_y$ to $\tilde{\omega}$, and we have
\begin{equation}
    \boldsymbol{\Phi}_0 = \left[
    \begin{array}{cc}
         \cos(c t_h)&  -\frac{c}{b}\sin({ct_h})\\
         \frac{b}{c}\sin(ct_h)& \cos(ct_h) 
    \end{array}
    \right]
\end{equation}
\begin{equation}
    \boldsymbol{\Phi}_1 = 
    \left[
    \begin{array}{cc}
         \frac{1}{c}\sin(c t_h)&  \frac{1}{b}\cos({ct_h})-\frac{1}{b}\\
         \frac{b}{c^2}-\frac{b}{c^2}\cos(ct_h)& \frac{1}{c}\sin(ct_h) 
    \end{array}
    \right]
\end{equation}
where 
\begin{equation}
b = \frac{I_z-I_x}{I_y}\omega_z,~~c = \sqrt{\frac{|(I_z-I_x)(I_y-I_z)|}{I_xI_y}}|\omega_z|
\end{equation}
Note that the detail expression of $\boldsymbol{\phi}$ varies with the quadrotor geometric property and the location of the failure rotor in the body frame.

From (\ref{eq:tilde_omega_th}), it is clear that $\tilde{\omega}$ is not only affected by the rotor generated moments, but also coupling moment (term $\boldsymbol{\omega}^B\times \boldsymbol{I}_v^B\boldsymbol{\omega}^B$ in (\ref{eq:M_dyn_equation})) as the function of initial angular velocity. Therefore, reducing $\tilde{\omega}$ is possible by leveraging these coupling moments after complete failure of rotors.

Consequently, the control allocation method constraining $\tilde{\omega}$ can be constructed as 
\begin{equation}
    \begin{array}{cccc}
\mathrm{P}2:
    &\underset{\boldsymbol{u},d}{\text{min}}
    &\left(\boldsymbol{\mu}_{\mathrm{des}}-\boldsymbol{G}\boldsymbol{u}\right)^T\boldsymbol{W} \left(\boldsymbol{\mu}_{\mathrm{des}}-\boldsymbol{G}\boldsymbol{u}\right) + \lambda \boldsymbol{u}^T\boldsymbol{u} + \gamma d^2\\
    &\text{s.t.} 
    & \boldsymbol{\phi}\boldsymbol{\Phi}_1\hat{\boldsymbol{G}}_{m}\boldsymbol{u}\leq -\boldsymbol{\phi}\boldsymbol{\Phi}_0[\omega_x,~\omega_y]^T + \Tilde{\omega}_\mathrm{max} + d\\
    & &  -d\leq0\\
    & &  \boldsymbol{u}_\mathrm{min} \leq \boldsymbol{u} \leq \boldsymbol{u}_\mathrm{max}\\
\end{array}
\label{eq:P2_optimization}
\end{equation}
where the first constraint stems from (\ref{eq:tilde_omega_th}), which sets limitations to the $\tilde{\omega}$ after $t_h$ by $\tilde{\omega}_\mathrm{max}$. The slack variable $d$ is added to guarantee the solution of above optimization problem; $\gamma>0$ is a weight added to the slack variable thereof. Note that the recovery performance is affected by three parameters: $t_h$, $\Tilde{\omega}_\mathrm{max}$ and $\gamma$. In general, the constraint of $\tilde{\omega}$ is more strict with a larger $t_h$,  $\gamma$ and a smaller $\Tilde{\omega}_\mathrm{max}$. P2 is a constrained quadratic programming problem which can be efficiently solved on-line using, for instance, the Active-set Algorithm \cite{Petersen2006} and the Interior Point Method~\cite{Vanderbei1999}.

After obtaining the reference thrust of each rotor by solving the quadratic programming problem P2, the RPM command or PWM command can be subsequently calculated using a model obtained by propeller static thrust tests, which is omitted in this paper for readability. 

\section{Simulation Validation}
\label{sec:simulation}
\subsection{Case Study: Comparison Between P1 and P2 Allocation}
% \subsection{Effect of Flight Envelope Protection P2}
\begin{figure}[t!]
    \centering
    \includegraphics[scale = 0.56]{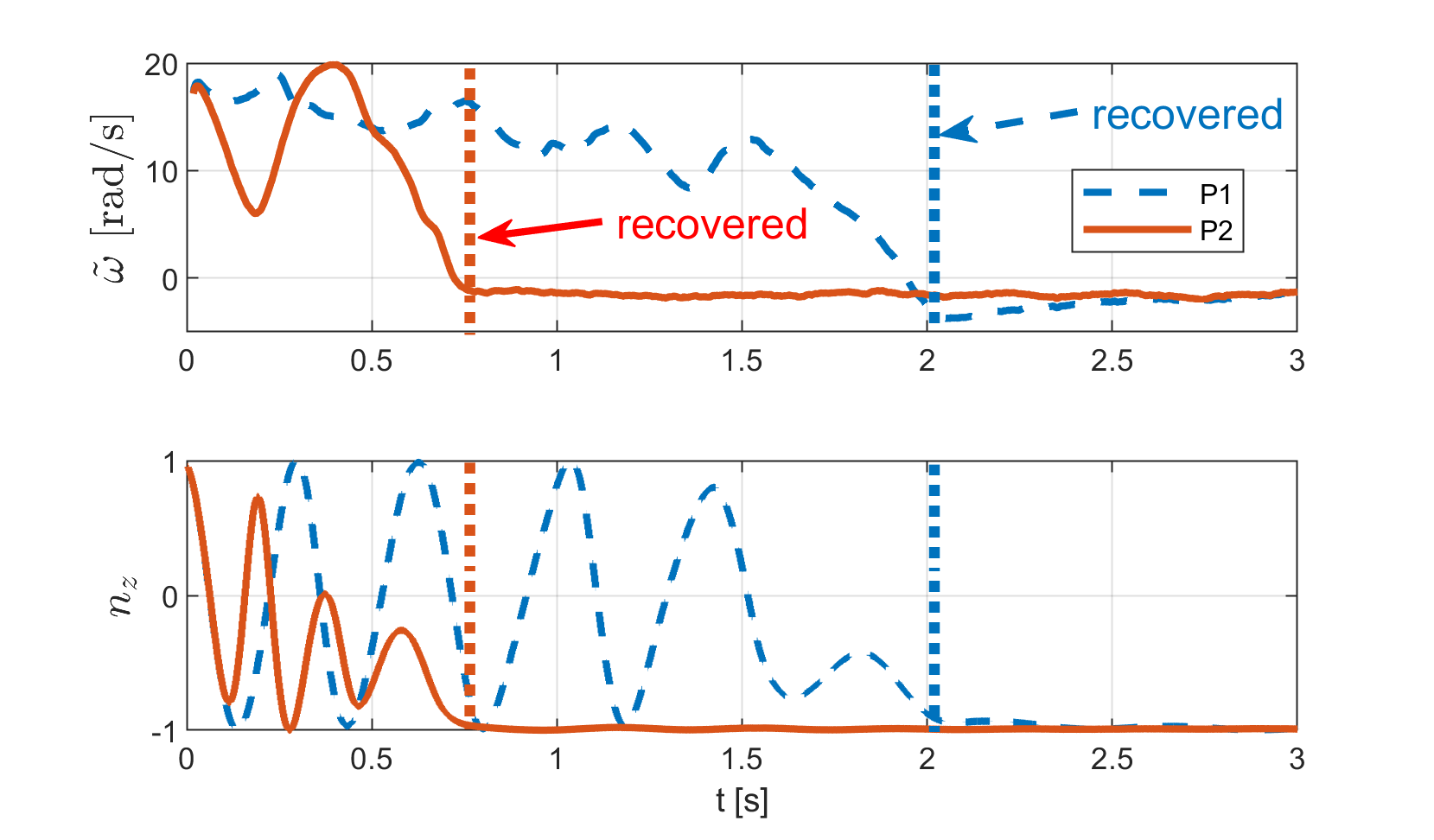}
    \caption{Two trajectories initialized from the same condition while using different allocation methods. The upper plot shows the angular rates about the unrecoverable axis $\tilde{\omega}$. The method P2 can more effectively suppress $\tilde{\omega}$ than P1. The lower plot shows the vertical component of the thrust direction $\boldsymbol{n}$, namely $n_z$, which should converge to -1 when the vehicle thrust vector points upward. It is clear the method P2 results in much faster recovery speed than P1.}
    \label{fig:recovery_P1vsP2}
\end{figure}
The proposed controller is first of all validated in a 6-DoF simulation. The simulation platform uses the quadrotor model developed in \cite{Sun2019}, which takes complex aerodynamic effects into account. The quadrotor inertial and geometric parameters are given in Table.~\ref{tab:Bebop2 parameters}.  One of the innovations proposed in this article is utilizing P2 from (\ref{eq:P2_optimization}) to replace P1 in (\ref{eq:P1_optimization}), such that the undesirable angular rate $\tilde{\omega}$ can be suppressed. Fig.~\ref{fig:recovery_P1vsP2} shows $\tilde{\omega}$ and $n_z$ of two recovery maneuvers using P1 and P2 as allocation methods respectively. In both simulations, the failure of rotor 4 occurs when $\boldsymbol{n} = [-0.2,~0.2,~0.98]^T$ and $\boldsymbol{\omega}^B = [-15,~15,~0]$. At this moment, the drone is almost upside down with a large $\tilde{\omega}$ at 17.3~rad/s. The target thrust orientation of both are set as $\boldsymbol{n}_\mathrm{des} = [0,~0,~-1]^T$, namely vertically upwards. It is clear that the trajectory with P2 allocation can effectively suppress $\tilde{\omega}$. Thereby the drone could recover its attitude within around 0.7~s, whereas the same problem without restraining $\tilde{\omega}$ recovers at around 2~s.
\subsection{Monte-Carlo Simulation}
% \begin{figure}
%     \centering
%     \includegraphics[scale=0.6]{figure/MC_sim_h_x_mix.png}
%     \caption{3D trajectory of a set of Monte-Carlo simulation including 100 flights initialized from random attitude and angular velocities. Subplot (a) shows trajectories using the benchmark method proposed by \cite{Sun2018} while subplot (b) shows trajectories using the proposed method.}
%     \label{fig:3D-MC}
% \end{figure}
\begin{figure}
    \centering
    \includegraphics[scale=0.60]{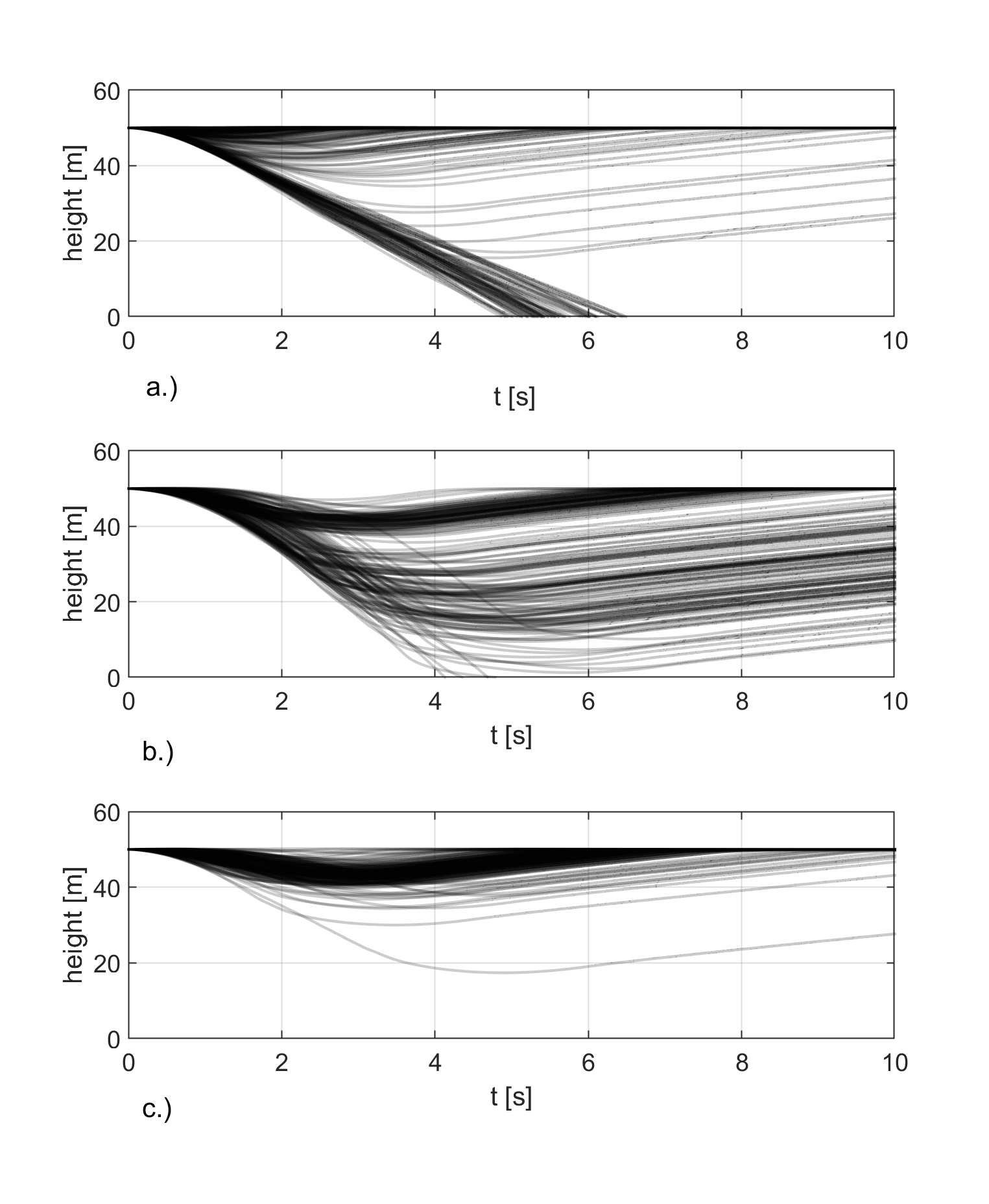}
    \caption{Altitude time series of a set of Monte-Carlo simulation including 200 flights initialized from random attitude and angular velocities with different flight control methods. a.) The benchmark method. b.) The proposed method but using P1 allocation. c.) The proposed method.}
    \label{fig:3D-MC}
\end{figure}

\begin{figure}
    \centering
    \includegraphics[scale=0.60]{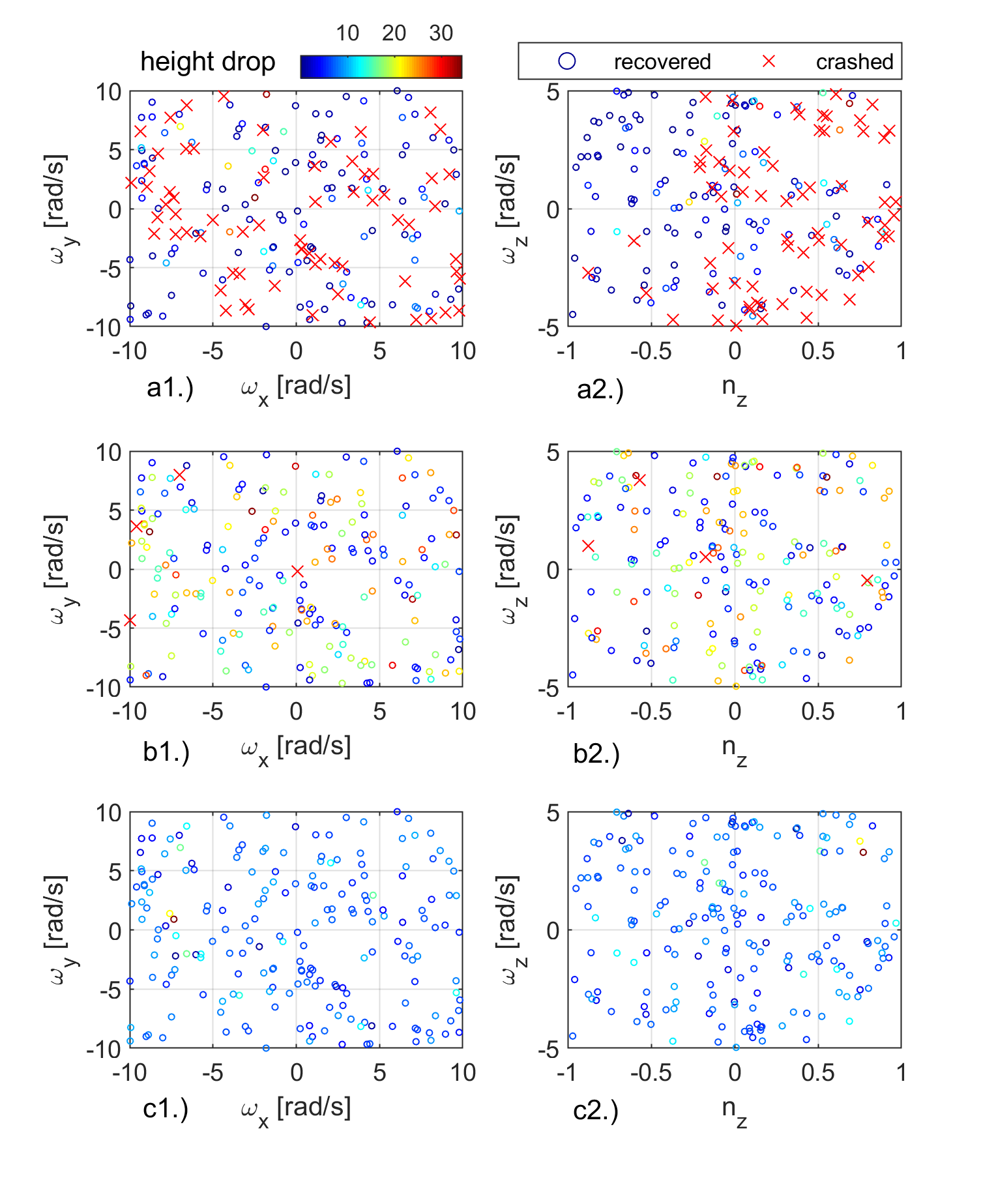}
    \caption{Scatter plot of the initial conditions of the Monte-Carlo simulation with colors showing the maximum height drop. The crashed flights are shown in red cross markers. a.) The benchmark method. b.) The proposed method but using P1 allocation. c.) The proposed method.}
    \label{fig:MC_initialcondition}
\end{figure}
A set of Monte-Carlo simulations are conducted to validate the proposed method. Another two methods are compared in these simulations: the method with P1 allocation, and the benchmark control method proposed by \cite{Sun2018}. For each method, 200 trajectories are simulated from different initial conditions.

We simulate the scenario where the failure of rotor 4 happens during the forward flight at speed. The initial position and velocity of these flights are set as $\boldsymbol{\xi}_0 = [0,~0,-50]^T$~m, $\boldsymbol{v}_0 = [10,~0,~0]^T$~m/s. The initial attitude is randomly selected in the entire SO(3), and the initial angular velocity $\boldsymbol{\omega}^B_0\sim U(-\boldsymbol{\omega}_\mathrm{0,max},~\boldsymbol{\omega}_\mathrm{0,max})$ where $\boldsymbol{\omega}_\mathrm{0,max}=[10,~10,~5]^T$~rad/s.

The altitude time series of different methods are plotted in Fig.~\ref{fig:3D-MC}. And Fig.~\ref{fig:MC_initialcondition} shows the scatter plot of the initial conditions of these three methods with color showing the maximum height drop. For the benchmark method, there are 67 out of 200 flights crashed. Most of these crashed flights marked in red crosses concentrate in the area with positive initial $n_z$ which indicate downward pointing initial thrust orientations (Fig.~\ref{fig:MC_initialcondition}-a2). On the other hand, the initial angular rates seem no special effect on the recovery performance. For method using P1 allocation shown in Fig.~\ref{fig:3D-MC}-b and Fig.~\ref{fig:MC_initialcondition}-b, there are 4 crashes but many of the rest recover after dropping for a large amount of altitude. There are two crashes concentrate on the top-left $\omega_x$-$\omega_y$ plane of Fig.~\ref{fig:MC_initialcondition}-b1 meaning that these flights are initialized with large $\tilde{\omega}$. In comparison, the proposed controller using P2 allocation method recovers the damaged drone in all of the 200 flights. 95\% of these flights could recover with a height drop of less than 10~m while only 1 flight recovers after dropping over 30~m, as is shown in Fig.~\ref{fig:3D-MC}-c and Fig.~\ref{fig:MC_initialcondition}-c. 

% Fig.~\ref{fig:MC_initialcondition} shows the scattered plot of the initial conditions of these three methods with color showing the height drop. For the benchmark method on the top two figures, most crashed flights marked in red crosses concentrate in the area with positive initial $n_z$ which indicates an upside down initial pose (Fig.~\ref{fig:MC_initialcondition}-a2). On the other hand, the initial angular rates seems no special effect on the recovery performance. For the method using P1 allocation shown in the mid figures, the height drops are in general larger than the proposed method. There are two crashes concentrate on the top-left $\omega_x$-$\omega_y$ plane meaning that these flights are initialized with large $\tilde{\omega}$.

\section{Experimental Validation}
\label{sec:experiment}
\begin{figure}
    \centering
    \includegraphics[scale=0.18]{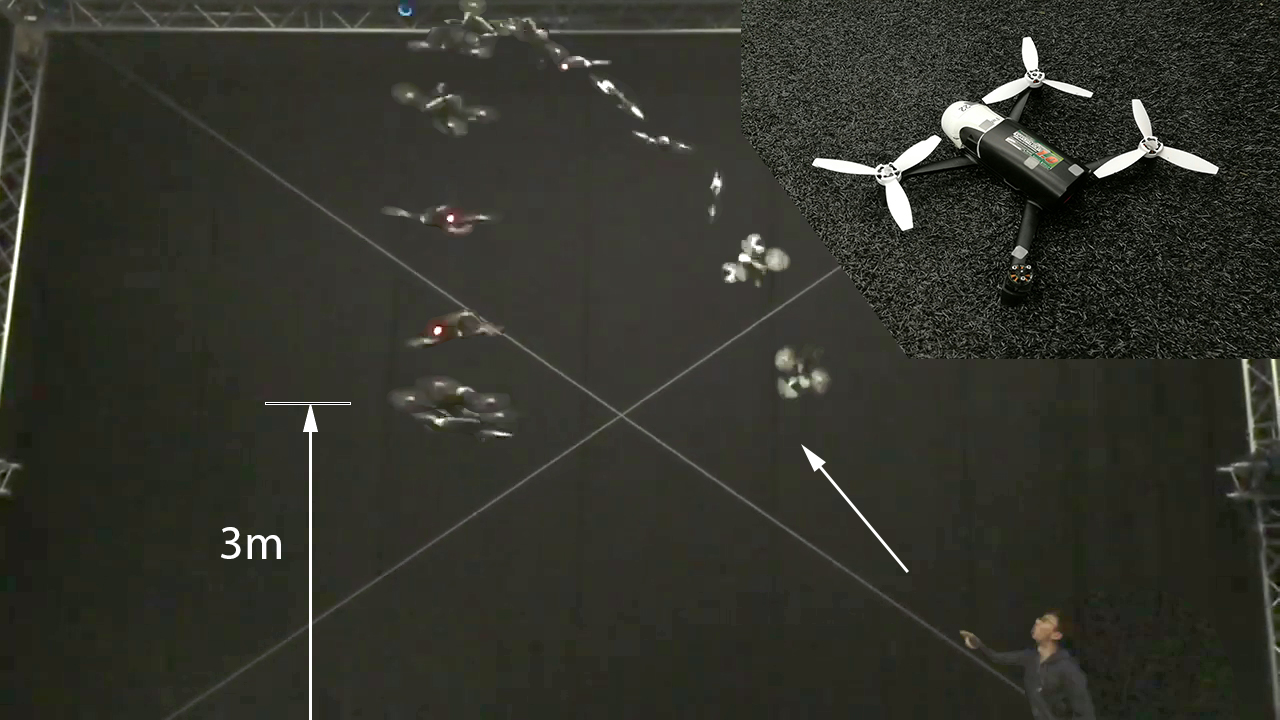}
    \caption{Snapshots of the quadrotor recovery maneuver after being tossed into the air. The drone was finally stabilized at 3m above the ground. Right top corner shows the photo of the tested quadrotor of which the left-back rotor was removed. (Video clip link: https://youtu.be/hrr2BzPLaMg)}
    \label{fig:time-lapse}
\end{figure}

The proposed method is also validated in the real flight environment. The tested platform is a modified Parrot Bebop 2 quadrotor, as Fig.~\ref{fig:time-lapse} shows. The parameters of this quadrotor are given in Table.~\ref{tab:Bebop2 parameters}. The flight was conducted in the Cyberzoo, TU Delft where 12 cameras from the motion capturing system (Optitrack) measured the position of 6 reflective markers attached to the drone in 120~Hz. The position information was then transmitted to the drone via WiFi, and the controller was run on-board in 500~Hz. The processor of the drone is a Parrot P7 dual-core CPU Cortex 9, and the IMU is MPU6050 for angular rate and specific force measurements.

\begin{figure}
    \centering
    \includegraphics[scale=0.64]{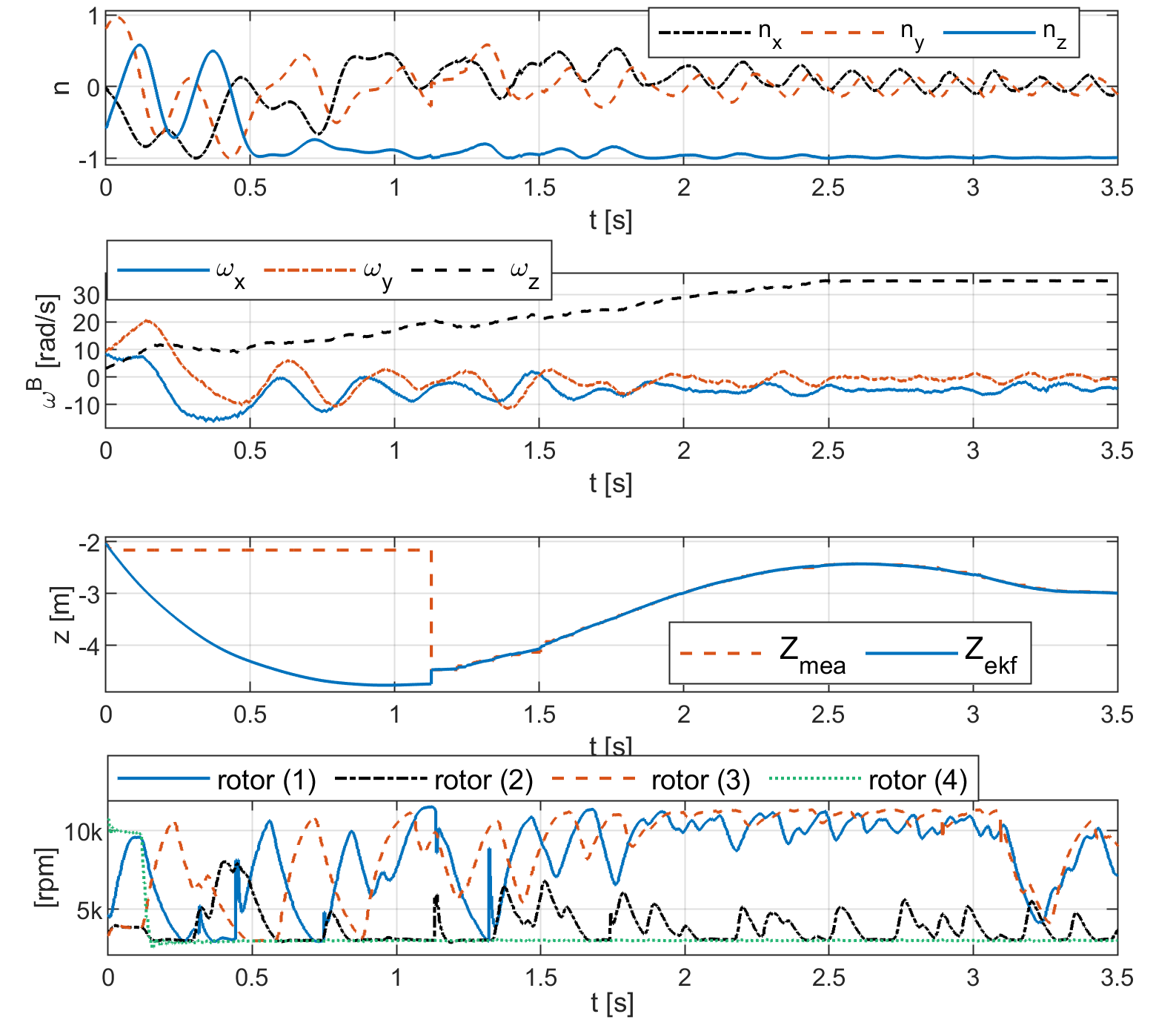}
    \caption{Time history of the recovery maneuver. Subfigures from top to bottom present the thrust orientation $\boldsymbol{n}$, angular rates $\omega^B$, altitude $z$ and rotor speeds respectively. }
    \label{fig:recover_experiments_data}
\end{figure}
To create the arbitrary initial condition, we threw the quadrotor with failure of rotor 4 into the air as Fig.~\ref{fig:time-lapse} shows. After reaching an altitude of 2 meters, the drone started recovering. Fig.~\ref{fig:recover_experiments_data} shows the reduced attitude $\boldsymbol{n}$, the angular rates, height and the rotor RPM in the recovery process. The drone was finally recovered and stayed at 3~m over the ground with a fast yaw rate. The controller parameters of this set of the test are listed in Table.~\ref{tab:control_gains}.
\begin{table}[t]
  \caption{Inertial and geometric properties of the tested quadrotor.}
  \begin{center}
  \begin{tabular}{c c c}
  \hline\hline
  parameter & value & unit\\
  \hline
   $\boldsymbol{I}^B_{v}$& diag(1.45,~1.26,~2.52)$\times 10^{-3}$& kgm$^2$ \\
   $m$, $l$, $\beta$ & 0.41, 0.145, 52.6 & kg, m, deg\\
   $s$, $\sigma$ & 1, 0.01 & -\\
   \hline
  \end{tabular}
  \label{tab:Bebop2 parameters}
  \end{center}
\end{table}  

\begin{table}[t]
  \caption{Control parameters of the real-life flight test.}
  \begin{center}
  \begin{tabular}{c c | c c}
  \hline\hline
  par. & value & par. & value \\
  \hline
   $\boldsymbol{K}_{p,\mathrm{pos}}$& diag$(1,1,15)$& $\boldsymbol{K}_{p,\mathrm{vel}}$ & diag$(2,2,25)$\\
   $\boldsymbol{K}_{i,\mathrm{vel}}$ & diag$(1,1,5)$& $(\theta_1,~\theta_2)$ & (30,~70) deg \\
   $k_{p,\mathrm{att}}$ & 8 & $\boldsymbol{K}_{p,\mathrm{rate}}$ & diag$(15,15,1)$  \\
   $\boldsymbol{W}$ & diag$(10^4,10^4,10^2,4)$ & $(\lambda,\gamma,t_h,\tilde{\omega}_\mathrm{max})$ & $(0.1,10^5,0.1,5)$  \\
   \hline
  \end{tabular}
  \label{tab:control_gains}
  \end{center}
\end{table}  
Since the motion capturing system is unable to measure the position of the drone with large attitude deviations from the hovering condition, an Extended Kalman Filter (EKF) is applied to fuse the camera measurements with the IMU measurements to obtain the position, velocity and attitude estimations. The 3rd subplot of Fig.~\ref{fig:recover_experiments_data} also shows EKF estimated altitude compared with the raw measurements and the latter keeps constant before $t=1.3$~s due to loss of tracking of the reflective markers.  

The in-door tests have a success rate of 71\% (46 out of 65 throws). However, those initialized from upside-down orientations and large $\tilde{\omega}$ is rather hard to recover before touching the ground. This is because of the height limitation of the laboratory (6 meters effective height) while it requires about 10 meters to completely recover from the upset condition. Therefore, out-door flight tests will be performed in future research, together with improved state estimation methods.

\section{Conclusions}
An upset recovery control method for a quadrotor with one rotor failure has been proposed and tested in this research. The controller can stabilize the quadrotor from arbitrary initial orientations and a wide range of angular velocities to the relaxed hovering condition. A novel control allocation approach is developed to suppress the undesirable angular velocities, which is important to the recovery performance. To demonstrate the reliability of the method, we have conducted Monte-Carlo simulations from random initial conditions. It has shown that the proposed method can timely recover the quadrotor with a height drop of less than 10~m in over 95\% flights. In the real-flight test, the controller can recover the damaged quadrotor after being randomly tossed into the air. Further tests in outdoor environments, with onboard state estimation, are suggested for future research.
\addtolength{\textheight}{-9cm}   % This command serves to balance the column lengths

%%%%%%%%%%%%%%%%%%%%%%%%%%%%%%%%%%%%%%%%%%%%%%%%%%%%%%%%%%%%%%%%%%%%%%%%%%%%%%%%

%%%%%%%%%%%%%%%%%%%%%%%%%%%%%%%%%%%%%%%%%%%%%%%%%%%%%%%%%%%%%%%%%%%%%%%%%%%%%%%%

% \label{sec:appendex}
% \subsection{Expression of $\boldsymbol{\Phi}_0$ and $\boldsymbol{\Phi}_1$}
% \begin{equation}
%     \boldsymbol{\Phi}_0 = \left[
%     \begin{array}{cc}
%          \cos(w t_h)&  -\sin({wt_h})\\
%          \frac{b}{w}\sin(wt_h)& \cos(wt_h) 
%     \end{array}
%     \right]
% \end{equation}
% \begin{equation}
%     \boldsymbol{\Phi}_1 = 
%     \left[
%     \begin{array}{cc}
%          \sin(w t_h)/w&  (\cos({wt_h})-1)/b\\
%          \frac{b}{w^2}(1-\cos(wt_h))& \sin(wt_h)/w 
%     \end{array}
%     \right]
% \end{equation}
% where 
% $$w = \sqrt{\frac{(I_z-I_x)(I_y-I_z)}{I_xI_y}}|\omega_z|,~~b = \frac{I_z-I_x}{I_y}\omega_z$$
\section*{Acknowledgement}
The authors would like to thank Xuerui Wang and the MAVLab for their support during flight tests.
%%%%%%%%%%%%%%%%%%%%%%%%%%%%%%%%%%%%%%%%%%%%%%%%%%%%%%%%%%%%%%%%%%%%%%%%%%%%%%%%

\bibliographystyle{ieeetr}
\bibliography{reference.bib}

\end{document}